# A Deep Learning Framework with Geographic Information Adaptive Loss for Remote Sensing Images based UAV Self-Positioning

**Mingkun Li[a,1,*], Ziming Wang[b,1], Guang Huo[c], Wei Chen[d], Xiaoning Zhao[e]**

[a] *China Academy of Aerospace Science and Innovation, No.2 Xinjiekouwai Street, Xicheng District, Beijing City*, mingkunli.96@gmail.com
[b] *China Academy of Aerospace Science and Innovation, No.2 Xinjiekouwai Street, Xicheng District, Beijing City*,, wangziming@lsec.cc.ac.cn
[c] *China Academy of Aerospace Science and Innovation, No.2 Xinjiekouwai Street, Xicheng District, Beijing City*, 2325373700@qq.com
[d] *China Academy of Aerospace Science and Innovation, No.2 Xinjiekouwai Street, Xicheng District, Beijing City*, chendavi_buaa@163.com
[e] *China Academy of Aerospace Science and Innovation, No.2 Xinjiekouwai Street, Xicheng District, Beijing City*, zz2050shiyanshi@163.com
* Corresponding Author
[1] Equal Contribution

## Abstract

With the expanding application scope of unmanned aerial vehicles (UAVs), the demand for stable UAV control has significantly increased. However, in complex environments, GPS signals are prone to interference, resulting in ineffective UAV positioning. Therefore, self-positioning of UAVs in GPS-denied environments has become a critical objective. Some methods obtain geolocation information in GPS-denied environments by matching ground objects in the UAV viewpoint with remote sensing images. However, most of these methods only provide coarse-level positioning, which satisfies cross-view geo-localization but cannot support precise UAV positioning tasks. Consequently, this paper focuses on a newer and more challenging task: precise UAV self-positioning based on remote sensing images. This approach not only considers the features of ground objects but also accounts for the spatial distribution of objects in the images. To address this challenge, we present a deep learning framework with geographic information adaptive loss, which achieves precise localization by aligning UAV images with corresponding satellite imagery in fine detail through the integration of geographic information from multiple perspectives. To validate the effectiveness of the proposed method, we conducted a series of experiments. The results demonstrate the method's efficacy in enabling UAVs to achieve precise self-positioning using remote sensing imagery.

**Acronyms/Abbreviations**
**UAV**: Unmanned Aerial Vehicle
**GNSS**: Global Navigation Satellite System
**SDM**: Spatial Distance Metric
**G²CL**: Geographic Information-Guided Contrastive Learning

## 1. Introduction

With the rapid expansion of UAV applications, the need for stable control has become increasingly critical, particularly in complex environments where GPS signals are prone to interference, leading to the loss of effective positioning. Ensuring reliable UAV self-positioning in GPS-denied environments has thus emerged as a significant challenge and a primary objective in UAV operations, which directly impacts the operational effectiveness and safety of UAVs [1,2,3].

In recent years, research has increasingly focused on cross-view geo-localization tasks, which involve matching UAV perspective images with remote sensing images to obtain geographic location information when GPS signals are unavailable [4,5,6]. However, most existing methods depend heavily on ground objects as reference points for cross-view matching [7,8]. While this approach can provide coarse localization sufficient for basic cross-view geo-localization requirements, it falls short in supporting the precise positioning tasks essential for many UAV operations [1]. This limitation becomes particularly evident in scenarios demanding fine-grained spatial accuracy, where even small errors can have significant operational consequences. For instance, while coarse localization may be adequate for general area navigation, it is insufficient for tasks requiring pinpoint accuracy, such as landing or navigating through cluttered environments. Recognizing this gap, our work aims to face a critical and hard challenge: UAV self-positioning based on remote sensing images[1,2]. Unlike cross-view geo-localization taret, UAV self-positioning demands not only the ability to perform coarse-grained matching of the UAV's current position across a large dataset of remote sensing images but also the capability to detect fine-grained spatial changes in objects across multiple nearby images. This dual requirement is critical for achieving precise distance perception and accurate positioning, essential for autonomous UAV navigation and task execution in GPS-denied environments. As shown in Figure 1, it can be observed that sampling points in cross-view geo-localization tasks are not continuous, with UAV images often capturing the same target from different angles. In contrast, datasets designed for UAV self-positioning focus more on continuous sampling of locations, leading to more prominent overlapping regions within the sampled images.

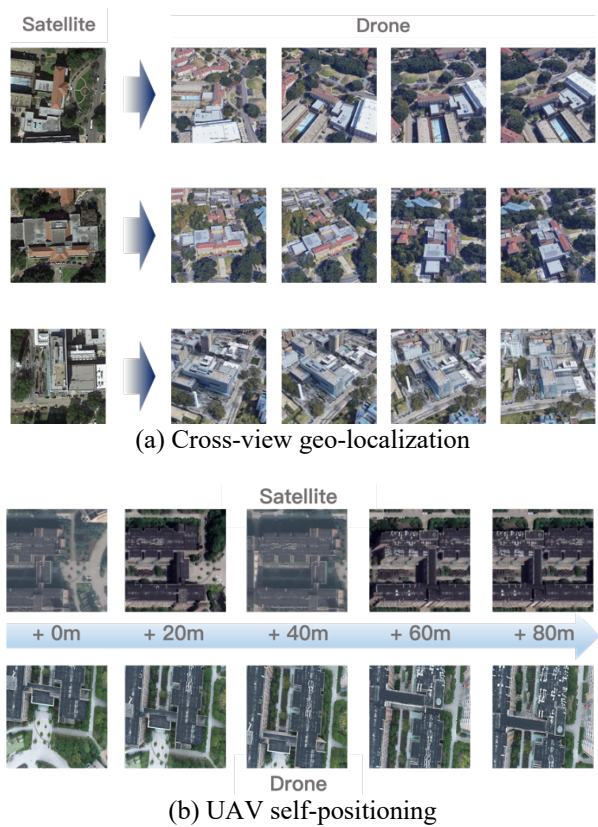

(a) Cross-view geo-localization

(b) UAV self-positioning

Fig. 2. Discrepancies in sample between cross-view geo-localization(a) and UAV self-positioning(b).

To address this challenge, we propose a simple yet effective deep learning framework specifically designed to meet the precise localization needs of UAVs in GPS-denied environments. First, we enhance the model's capability to extract distinguishable features from different perspectives by aligning UAV images with corresponding satellite images as input pairs. Specifically, we pair satellite images with UAV images captured at the same GPS location, treating these pairs as positive samples, while treating pairs from different locations as negative samples. We then employ the InfoNCE [4] loss function to capture consistency information between these image pairs, enabling the model to learn coarse-grained consistency features. This is achieved by emphasizing the similarity within positive sample pairs and maximizing the separation of negative sample pairs in the feature space. These features are crucial at the initial matching level, where the model must identify potential candidate regions within the remote sensing images that correspond to the UAV's view.

Building on this foundation, we introduce an adaptive loss function that incorporates geographic auxiliary information. Combined with our positive and negative sample construction strategy, this adaptive loss function enables the model to adjust the feature space distance between positive and negative sample pairs based on GPS location data. This adjustment ensures that the distribution of image samples in the feature space closely reflects their actual geographic distribution in the physical world. This fine-grained alignment in the feature space is particularly crucial for refining the model's ability to perceive and differentiate subtle spatial variations. As a result, our approach allows the model to extract highly discriminative features from consecutive images, thereby facilitating fine-grained and precise UAV self-positioning.

To validate the effectiveness of our proposed method, we conducted extensive tests using the DenseUAV[1] dataset, which is specifically designed for UAV precise localization tasks. Remarkably, our method achieved superior results even when using a smaller amount of data, demonstrating the robustness and efficiency of our approach. These experimental results confirm that our method effectively supports UAV precise self-localization using remote sensing images, even under data-constrained conditions. This not only highlights the practicality of our approach but also underscores its potential for deployment in real-world scenarios where data availability may be limited.

Our contributions are summarized as follows.

1. We introduced a novel contrastive learning framework tailored specifically for the UAV self-localization task. This framework aligns UAV images with corresponding satellite images, enabling the model to effectively learn and extract discriminative features necessary for accurate localization.

2. We proposed an adaptive loss function that integrates geographic auxiliary information. This method dynamically adjusts the feature space distance between positive and negative sample pairs, ensuring the model's learned feature distribution accurately reflects the spatial relationships in the real world.

3. Our method demonstrated robust performance on the DenseUAV dataset, designed for UAV precise localization tasks. Even with a reduced amount of data, our approach outperformed existing methods, validating its effectiveness in supporting UAV precise self-localization.

## 2. Related Work

### 2.1 Cross-View Geo-Localization

The cross-view geo-localization task focuses on matching images captured from different platforms, such as satellites, UAVs, and ground cameras, across varying altitudes[9,10,11]. This approach leverages the geographic coordinates embedded in satellite images to determine the location of the current view by retrieving the corresponding satellite image in the absence of Global Navigation Satellite System (GNSS) information. Initially, research in cross-view geo-localization concentrated on aligning ground camera perspectives with satellite views to provide location information for applications such as ground vehicle navigation.

The primary challenge in cross-view geo-localization is extracting consistent feature information for matching images captured from different perspectives [4,12]. With the advancement of deep learning, the superior feature extraction capabilities of deep neural networks have garnered significant attention. In the context of cross-view geo-localization, an increasing number of methods are utilizing contrastive learning to train models to learn the consistency between satellite images and corresponding ground images, thereby extracting invariant features. For instance, SAFA[13] employs a two-step approach, aligning aerial and ground panoramic images using polar transforms and incorporating spatial attention mechanisms for feature aggregation, which enhances feature representation capabilities. [14] examined the robustness of aligning ground-view and aerial-view images by utilizing Siamese networks alongside contrastive learning. Despite these advancements, the nearly vertical viewpoints between ground and satellite platforms continue to present challenges in extracting consistent information solely from images.

As UAV technology continues to mature, an increasing number of scholars are focusing on utilizing UAV perspectives for target localization and tracking. Compared to ground platforms, UAVs offer a wider field of view and greater mobility, making them particularly well-suited for cross-view geo-localization tasks in conjunction with satellite platforms. Unlike ground-to-satellite localization, UAV-to-satellite localization benefits from reduced viewpoint variations and richer consistency information, resulting in higher accuracy within similar model frameworks. This has led to significant research interest[4,15], with numerous methods and datasets being developed to enhance the performance of UAV-to-satellite geo-localization.

One notable contribution is the University-1652[16] dataset, designed to facilitate the matching of UAV perspectives with satellite images for global UAV localization or target discovery. Numerous methods have been developed using this dataset as a foundation [17,18,19]. For example, Sample4-Geo[4] adopts a simple yet effective architecture that leverages contrastive learning with symmetric InfoNCE loss, achieving state-of-the-art results without relying on aggregation modules or complex pre-processing steps, thus improving generalization across different regions and achieve high performance on both ground-to-satellite and aerial-to-satellite targets.

However, traditional methods for UAV localization primarily focus on matching objects within the UAV's perspective, often overlooking the spatial discrepancies caused by the UAV's camera angle and the varying distances of objects from the center of the field of view. As a result, these methods are not directly applicable to precise UAV localization[1,3], necessitating further refinement and adaptation to address these challenges. In this paper, we focus on advancing UAV self-positioning by developing a model capable not only of locating objects but also of distinguishing fine-grained differences in geographic distance, thereby enhancing the accuracy and reliability of UAV autonomous localization.

### 2.2 UAV Self-positioning

To further address the challenges of UAV autonomous localization in the absence of GNSS support, recent research has shifted its focus from merely matching targets within the UAV's field of view to determining the UAV's relative position in space. This shift reflects the need for more precise localization that goes beyond simple image matching. In response, some researchers[1,2,3] have introduced innovative methods and corresponding datasets to tackle this problem.

One such contribution is the DenseUAV[1] dataset, which emphasizes dense sampling of real spatial images. This approach better captures the continuous perspective changes experienced during UAV flight, offering a more accurate representation of the UAV's changing viewpoint compared to previous datasets that relied on sparse sampling. Additionally, DenseUAV avoids using panoramic images for UAV view images, opting instead to position the UAV camera vertically towards the ground. This design choice mitigates the issue of viewpoint bias, which is common in datasets that include panoramic views.

Moreover, DenseUAV introduces a novel metric, the Spatial Distance Metric (SDM@K), specifically tailored for UAV autonomous localization tasks. Unlike traditional metrics that focus on one-to-one target matching accuracy, SDM@K prioritizes the relative position of the retrieved image in relation to the target point. This shift in focus aligns with the goal of improving UAV localization by emphasizing spatial relationships over individual object matching.

Building on this research, the present study proposes an adaptive loss function that integrates geographic auxiliary information to further enhance UAV autonomous localization. By incorporating geographic localization data into the model training process, this approach enables the model to place greater emphasis on relative spatial position information, thereby improving its ability to accurately perceive distances and achieve precise localization.

## 3. Method

In this section, we introduce a contrastive learning framework specifically designed to address the UAV self-positioning challenge by incorporating geographic auxiliary information. This approach, termed Geographic Information Guided Contrastive Learning (G2CL), enhances the model's ability to accurately determine UAV positions by integrating spatial data with traditional contrastive learning techniques.

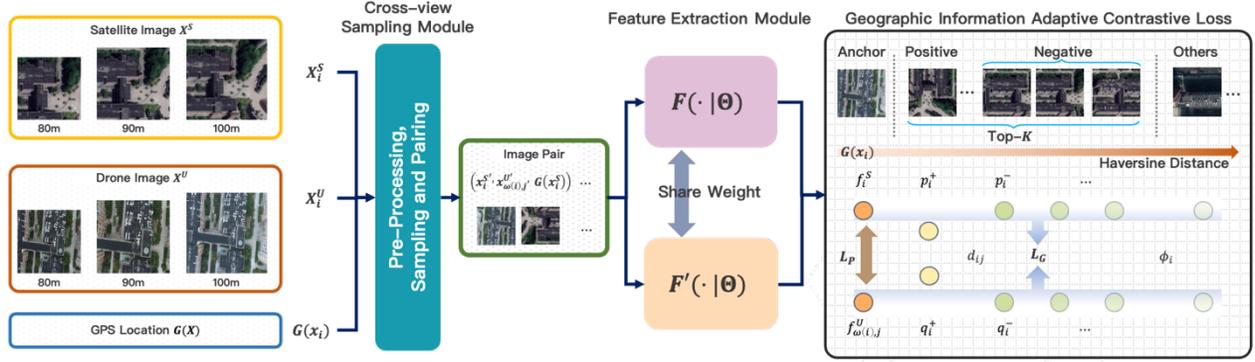

Fig. 2. The total framework of our proposed G$^2$CL.

### 3.1 Training Data Processing and Sampling Strategy

For clarity, we illustrate the complete framework of G2CL in Figure 1, which comprises three main components: a cross-view sampling module, a feature extraction module, and an adaptive loss function that utilizes auxiliary information. During the training stage, the cross-view sampling module pre-processes input samples, constructing pairs of UAV and satellite images based on geographic information. These paired samples are then processed by the feature extraction module, which leverages contrastive learning to extract invariant features between the paired UAV and satellite images, enabling coarse-grained cross-view matching. To effectively explore hidden spatial distance features at a fine-grained level, we introduce an adaptive loss function that integrates global localization information embedded in the images. This function adjusts the distance relationships between different samples in the feature space according to the localization data, allowing for the fine-grained differentiation of densely sampled spatial samples and achieving precise localization in such environments.

In Section 3.1, we present the basic information of the data used, the pre-processing methods, and the pairwise sampling strategy. In Section 3.2, we detail the contrastive learning framework employed for feature extraction. In Section 3.3, we describe the loss function used to train the contrastive learning framework, and in Section 3.4, we outline the detailed strategies for the training and testing phases.

For an UAV self-localization task, the training dataset $X = \{x_i\}_{i=1}^{L}$ with a total of $L$ images generally consists of two parts: remote sensing images from satellite perspectives $X^S = \{x_i^S\}_{i=1}^{M}$ with $M$ images, and UAV perspective image $X^U = \{x_i^U\}_{i=1}^{N}$, with $N$ images, where $M + N = L$. In the training dataset, for a satellite image $x_i^S$, it has a corresponding set of UAV perspective images with a total of $n$ images, and defined as $\{x_{\omega(i),1}^U, x_{\omega(i),2}^U, \dots, x_{\omega(i),n}^U\}$, where $\omega(i)$ is the index set of UAV images corresponding to $x_i^S$, and $(\omega(i), j)$ is the $j$-th image in the corresponding set $\omega(i)$. All images in the corresponding set share the same image center GPS location $G(x_i^S)$ with $x_i^S$, where $G(x_i)$ denotes the GPS location of image $x_i$ for both UAV and satellite perspectives. Furthermore, due to the dense sampling strategy, and in contrast to typical cross-view geo-localization datasets, the gallery for a satellite image $x_i^S$ includes not only UAV images $x_{\omega(i),j}^U$ with the same GPS location but also a few images that are close to the center point of the satellite image $x_i^S$, as well as images from distant geographical locations.

Before inputting into the G2CL framework, all images in the dataset $X$ undergo a pre-processing stage. Specifically, images from both satellite and UAV perspectives are first resized to a uniform size. Then, with a certain probability, each image is subjected to a data augmentation process, including operations like random flipping and color-jittering. We define the data augmentation operation as $\varphi(\cdot)$, and the augmented image as $x' = \varphi(x)$.

After generating the augmented image dataset $X' = \{x_1', x_2' \dots x_N'\}$, a sampling strategy is applied to construct input image pairs based on GPS location information. Specifically, for a satellite image $x_i^{S'}$, we pair it with images from the corresponding UAV perspective set $\{x_{\omega(i),j}^{U'}\}_{j=1}^n$, generating $n$ pairs that contain the corresponding GPS location information, such as $\{(x_i^{S'}, x_{\omega(i),j}^{U'}, G(x_i^S))\}_{j=1}^n$.

Thus, after the data pre-processing and sampling operations, the total input to G2CL is defined as $\sum_{i=1}^M \sum_{j=1}^{\Omega(i)} \{(x_i^{S'}, x_{\omega(i),j}^{U'}, G(x_i^S))\}$, where $\Omega(i)$ denotes the scale of the corresponding UAV perspective image set for satellite image $x_i^S$.

This design permits the network to process paired images as input samples. To effectively extract features from these samples, we propose a dual-branch network with a contribution parameter, which is used to simultaneously extract features from the paired images. This approach facilitates contrastive learning between features from different perspectives of the same location.

### 3.2 Geographic Information Guided Contrastive Learning Framework

In order to achieve an effective matching of images from both UAV and satellite perspectives, we propose the implementation of a contrastive learning framework. This framework is constituted by a dual-branch network with shared parameters, which is designed to simultaneously extract features from paired images. During the training phase, we employ the geo-information-based pairing method proposed in Section 3.1, whereby two images from different perspective an at the same geographical location are simultaneously input.

Specifically, as shown in Fig.1, our Geographic information Guided Contrastive Learning (G$^2$CL) framework is a siamese network, which composed of two branches $F(\cdot | \Theta)$ and $F'(\cdot | \Theta)$ with sharing parameter $\Theta$. Before training, we use the pretrained ConvNext-tiny[20] as the backbone and initialize the parameter $\Theta$ with pretrained dataset. During the training stage, both branches $F(\cdot | \Theta)$ and $F'(\cdot | \Theta)$ simultaneously extract different image features in the image pair. Specifically, given a image pair $(x_i^{S'}, x_{\omega(i),j}^{U'}, G(x_i^S))$, the first branch is used to extract the satellite perspective image feature $f_i^S$ and the second branch is used to extract the UAV perspective image feature $f_{\omega(i),j}^U$, where $f_i^S = F(x_i^{S'} | \Theta)$ and $f_{\omega(i),j}^U = F'(x_{\omega(i),j}^{U'} | \Theta)$, respectively. Then the features of satellite image $f_i^S$ with its corresponding image feature $f_{\omega(i),j}^U$ will be used to calculated the geographic adaptive loss value and update the parameters $\Theta$ as described in detail in Section 3.3.

The design of the two-branch network structure with shared parameters allows for the simultaneous extraction of image features from two different viewpoints within the same geographic location. Through combining with the geographic information adaptive loss function proposed in Section 3.3, this framework facilitates the mining of consistent information from the two viewpoints, thereby aligning the images from different perspectives and enabling their effective retrieval.

As illustrated in Fig. 1, our Geographic Information Guided Contrastive Learning (G2CL) framework employs a Siamese network composed of two branches, $F(\cdot | \Theta)$ and $F'(\cdot | \Theta)$, which share the parameter set $\Theta$. We utilize the pretrained ConvNext-tiny[20] model as the backbone and initialize the parameters $\Theta$ with a pretrained dataset. During the training stage, both branches $F(\cdot | \Theta)$ and $F'(\cdot | \Theta)$, simultaneously extract features from the images in each pair. Specifically, given an image pair $(x_i^{S'}, x_{\omega(i),j}^{U'}, G(x_i^S))$, the first branch extracts the satellite perspective image feature $f_i^S$, while the second branch extracts the UAV perspective image feature $f_{\omega(i),j}^U$, where $f_i^S = F(x_i^{S'} | \Theta)$ and $f_{\omega(i),j}^U = F'(x_{\omega(i),j}^{U'} | \Theta)$. The features $f_i^S$ and $f_{\omega(i),j}^U$ are then used to calculate the geographic adaptive loss, which updates the parameters $\Theta$ as detailed in Section 3.3.

The design of this two-branch network structure with shared parameters enables the simultaneous extraction of image features from two different perspectives within the same geographic location. Combined with the geographic information adaptive loss function described in Section 3.3, this framework effectively mines consistent information from the two viewpoints, aligns images from different perspectives, and enhances their retrieval accuracy.

### 3.3 Geographic Information Adaptive Contrastive Loss

To effectively encourage the model to learn consistent information between cross-platform perspectives, we have proposed a pairwise sampling method based on geographical information, along with a Siamese network structure with shared parameters that accepts paired samples as input. Building on this foundation, we introduce a contrastive loss function that incorporates geographical information, termed Geographic Information Adaptive Contrastive Loss. This loss function uses geographical information as a reference to control the distance between samples with different geographical distributions in the feature space, enabling adaptive adjustment based on the sample features.

#### 3.3.1    Paired-Level Contrastive Loss

Specifically, in G²CL, the geographic information adaptative contrastive loss contain two parts, paired-level contrastive loss $L_P$ and geographic-based adaptive loss $L_G$. For an image pair $(x_i^{S'}, x_{\omega(i),j}^{U'}, G(x_i^S))$ within a batch, the pair-level contrastive loss is defined as :

$$L_P\left(x_i^{S'}, x_{\omega(i),j}^{U'}\right) = -\log \frac{\exp\left(f_i^S \cdot f_{\omega(i),j}^U / \tau\right)}{\sum_{b=1}^B \exp\left(f_i^S \cdot f_b^U / \tau\right)}, \quad (1)$$

where $B$ is the batchsize of current input, and $\tau$ is the temperature parameters that controls the scaling of the feature similarity value. The term $f_b^U$ denotes the UAV perspective images within the current batch, including both paired image and non-paired images relative to $f_i^S$. The loss defined in Eq.(1) encourages the model to strengthen the correlation between features $f_i^S$ and $f_{\omega(i),j}^U$ of different platforms situated in the same geographical location, which facilitates invariance information extraction across cross-platform perspectives. Since the consistency information is derived from the paired images $(x_i^{S'}, x_{\omega(i),j}^{U'})$ following the sampling process, we refer to this component as the pair-level contrastive loss.

The contrast loss function at the paired-level encourages the model to learn consistent information about the same object from different perspectives, thereby effectively supporting the model to perform cross-perspective retrieval based on objects within the platform's field of view. This kind of "coarse-grained level" positioning based on objects is sufficient for cross-view geo-localization tasks, but it is inadequate for supporting the UAV self-localisation task of the drone. In the case of dense sampling strategy, the same object may appear frequently in the field of view of sampling points at similar distances. Therefore, relying solely on the object for UAV self-positioning is insufficient and the model must also learn the relative position information between images with near location. To solve this problem, in this paper we propose a geographic-based adaptive loss $L_G$.

### 3.3.2    Geographic-based Adaptive Loss

In order to enable the model to distinguish between image samples that are geographically neighbouring, we attempt to ensure that the distribution of sample features is relatively consistent with their geographical distribution. This implies that if the points in the geographical space are in close proximity, they are also proximate in the feature space. Based on this concept, we initially delineate the strategy for defining the neighbour set in both the feature space and the geographical space. Subsequently, we utilise the constructed neighbourhood information to calculate the geographical loss function corresponding to the various samples, based on the relative distribution of the extracted features from the current model and the neighbourhood. This enables us to achieve an adaptive loss value for the different samples.

Firstly, we defined the neighbour select metric method for geographical space and feature space, respectively. In the feature space, we use the Euclidean Distance to metric the feature distance $d_{i,j}$ as follows:

$$d_{i,j} = ||f_i - f_j||, \quad (2)$$

where $f_i$ and $f_j$ is the feature for image $x_i$ and $x_j$. In the geographical space, we use the haversine distance $h_{i,j}$ to calculate the sampling geographical distance, and use the top $k$ nearest neighbour to construct the geographical neighbourhood set, defined as $\Phi = \{\Phi_1, \Phi_2 \dots \Phi_N\}$ and each $\Phi_i$ contain $k$ neighbour indexes under distance increasing sort.

Once the neighbourhood set has been constructed, we put forward a geographic-based adaptive loss function to quantify the discrepancy between the prevailing feature distribution and the actual geographical location. Specifically, for one image pair, the total geographic-based adaptive loss is comprised of three constituent parts: the satellite part loss $L_{GS}$, the UAV part loss $L_{GU}$ and the cross-view part loss $L_{GC}$, which is defined as:

$$L_G = L_{GS} + L_{GU} + L_{GC}, \quad (3)$$

where $L_{GS}$ is the satellite perspective loss for a satellite image $x_i^S$ in current batch and defined as:

$$L_{GS}(x_i^S) = \log\left(1 + exp\left(\phi_i(p_i^+ - p_i^-)\right)\right), \quad (4)$$

$p_i^+$ and $p_i^-$ is the positive and negative sample of satellite image within the same perspective in current batch in the feature space. The positive sample $p_i^+$ and negative samples $p_i^-$ are selected by the hard example mining strategy.

Specifically, in the case of the current batch, the hard example mining strategy selects different GPS location sample with minimum feature distance as the negative sample, and the same GPS location sample with maximum feature distance is the positive sample, which defined as:

$$p_i^+ = \max_{G(x_i)=G(x_j)} d_{i,j}, \;\; p_i^- = \min_{G(x_i)\neq G(x_j)} d_{i,j}, \quad (5)$$

where $d_{i,j}$ is the feature distance as defined in Eq.(2), and $\phi_i$ is the adaptative weight parameter for $L_G$, defined as follows:

$$\phi_i = \begin{cases} \alpha, & if \;\; \omega(p_i^-) \notin \Phi_i \\ \alpha/||h_i|| & if \;\; \omega(p_i^-) \in \Phi_i \end{cases}. \quad (6)$$

In Eq.(6), $\omega(p_i^-)$ is the image index of $p_i^-$, and $\|h_i\|$ is the neighbour normalised metric of sample $x_i$, which is defined as :

$$\|h_i\| = h_{i,\omega(p_i^-)} / \min_{j \in \Phi_i} h_{i,j}, \qquad (7)$$

where $h_{i,\omega(p_i^-)}$ is the haversine distance between $x_i$ and $x_{\omega(p_i^-)}$, $\Phi_i$ is the geographical neighbour set of $x_i$. $\alpha$ is a hyperparameter in G²CL and we will detailed the effect of it in the experiment section.

Based on Eq.(4), for the satellite perspective, the model will adaptively adjust the weight parameters in consideration of the presence of geographical neighbours within the current batch and the geographical distance between samples. The equation penalises the distance between the anchor point and the positive sample, relative to the distance between the anchor point and the negative sample. This encourages the model to maximise the distance between negative samples and minimise the distance between positive samples. For samples in the neighbourhood set, the model will further increase the corresponding weight parameters, thereby forcing their distribution in the feature space to align with the geographical distribution.

The UAV part loss $L_{GU}$ and the cross-view part loss $L_{GC}$ are defined as follow:

$$L_{GU}\left(x_{\omega(i),j}^U\right) = \log\left(1 + exp\left(\phi_{\omega(i),j}\left(q_{\omega(i),j}^+ - q_{\omega(i),j}^-\right)\right)\right), (8)$$

$$L_{GC}\left(x_i^S, x_{\omega(i),j}^U\right) = \log(1 + exp(\phi_i(z_i^+ - z_i^-))), \qquad (9)$$

In Eq.(8), $q_{\omega(i),j}^+$ and $q_{\omega(i),j}^-$ is the positive and negative samples of UAV perspective $x_{\omega(i),j}^U$ in current batch, $z_i^+$ is the cross-view feature positive samples, which calculate the feature distance between positive and negative for $f_i^C$, and $f_i^C$ is generated by the feature dim-concat operation on $f_i^S$ and $f_{\omega(i),j}^U$:

$$f_i^C = concat\left(f_i^S, f_{\omega(i),j}^U, axis = 0\right), \qquad (10)$$

and $\phi_{\omega(i),j}$ is the weight parameter which is calculated in the same manner as Eq (6).

Thus, The total loss of our G²CL framework is defined as :

$$L_{ALL} = L_p + L_G, \qquad (11)$$

The loss function Eq.(11) enables the model to identify and utilise consistency information between image pairs from different platform, while maintaining the sample features in accordance with their geographical distribution. In the section dedicated to experimental procedures, a series of experiments were conducted to demonstrate the efficacy of the method in accurately recognising spatial relative positions under a denser point distribution.

### 3.4 Inference Produce of G²CL

In the training phase, both $F(\cdot | \Theta)$ and $F'(\cdot | \Theta)$ are employed to extract features and update the parameter using AdamW[21]. In contrast, in the reference phase, only $F(\cdot | \Theta)$ is utilised for image extraction.

Since the objective of the drone self-positioning task is to align the drone's current perspective with that of the pre-existing satellite images, for UAV self-positioning dataset, the test part dataset contain both query image $X^Q = \{x_1^Q, x_2^Q, \ldots, x_n^Q\}$ and gallery image $X^G = \{x_1^G, x_2^G, \ldots, x_m^G\}$ from UAV and satellite platform, respectively. In the inference stage, the branch $F(\cdot | \Theta)$ extract both query and gallery image feature and match each image in query with gallery dataset. In particular, for image $x_i^Q$, we compute its Euclidean Distance in feature space with all images in $X^G$, and then sorted in ascending order. The evaluation criteria for the ranking results will be described in detail in the following experiment section.

## 4. Experiment and Results

In this section, we firstly describe the dataset and the evaluation metrics designed for the UAV self-positioning target, then we introduce the implementation details of parameter settings and training strategy of G²CL, then we provide a detail experiment sets to improve the effective of our proposed method including compare with other methods and sets of detailed ablation studies. Finally we make more evaluation based on visualization experiments and so on for a further analysis.

### 4.1 Dataset Description and Evaluation Metrics

In this subsection, we introduce the used dataset in experiments and the evaluation metric of UAV self-positioning target.

#### 4.1.1 Dataset Description

To evaluate the effectiveness of G²CL, we test our methods on DenseUAV, a dataset proposed specifically for the task of self-positioning of UAV. The dataset contains images collected by satellites and UAV platforms at different times and altitudes. Unlike conventional cross-view geolocation tasks, the dataset uses a dense sampling strategy. Specifically, for the UAV platform, the UAV perspective image is sampled at three altitudes of 80m, 90m and 100m, and at the same time, to meet the need for intensive sampling of the area, the UAV is continuously sampled at 20m intervals. For each location, DenseUAV collects corresponding satellite remote sensing images at three different scales in two time periods, 2020 and 2022. In total, the DenseUAV training set consists of 6768 UAV view images and

13536 satellite view images from 2256 sampling points. The query dataset consists of 2331 UAV images and 4662 satellite images from 777 sample points in four schools.

### 4.1.2 Evaluation Metrics

For the UAV self-positioning objective, following DenseUAV, we use both Recall@K and SDM@K to evaluate the model performance. Recall@K is widely used in the cross-view geo-localisation objective to measure the accuracy of the matching image sorting, and we use Racall@1 in the experiments. SDM@K is a metric proposed by DenseUAV for the specific purpose of self-localisation of UAVs. In contrast to Recall@K, which solely assesses the presence of the correct image within the results of the $K$ image sequences retrieved, SDM@K additionally evaluates the conformity of the sequence distribution of the remaining retrieved images to the corresponding geographical coordinate positioning distribution. In particular, for image retrieval results comprising a length of $K$, the SDM@K indicator exhibits a higher value when the first image is the correct match and the remaining $K$-1 images are selected from the gallery in accordance with their location proximity to the matching point.

### 4.2 Implementation Details

In our G²CL approach, we employ the ConvNext-Tiny[20] framework as the underlying foundation for both branches, and use the AdamW[21] optimizer with a batch size of 128, and the temperature parameter $\tau$ is set to 0.1 in Eq. (1).

Prior to training, the haversine distance between different sampling locations is calculated. During the training stage, the network is trained for a total of 50 epochs with an initial learning rate of 0.0001.

### 4.3 Comparison to the State-of-the-art Method

In this section, we present a comprehensive comparison with DenseUAV. Specifically, for a single sampling point, the following settings are used to evaluate model performance: (1-3) both 2020 and 2022 satellite images with a single scale, along with all UAV images, resulting in six image pairs per location; (4) 2022 satellite images at all scales, combined with all UAV images, resulting in nine image pairs per location; (5) all scales and timeframes of satellite images, together with all UAV images, yielding a total of eighteen image pairs per location; and (6) only the largest-scale 2022 satellite image and all UAV images, resulting in three image pairs per location.

Table 1 presents the performance of G²CL under each setting, alongside the corresponding performance reported by DenseUAV for settings (1-5).

Table 1. Comparison results on DenseUAV

| Method | | Satellite-View | | | | | Recall @1 | SDM @1 | SDM @3 | SDM @5 | SDM @10 |
| | | | Scale | | Time | | | | | | |
| | | Small | Middle | Big | 2020 | 2022 | | | | | |
| Dense UAV[1] | 1 | √ | | | √ | √ | 76.36 | 81.42 | 66.90 | 55.79 | 40.26 |
| | 2 | | √ | | √ | √ | 79.28 | 83.41 | 68.36 | 56.81 | 40.90 |
| | 3 | | | √ | √ | √ | 76.49 | 81.55 | 66.33 | 55.12 | 93.53 |
| | 4 | √ | √ | √ | | √ | 70.48 | 75.79 | 73.38 | 64.59 | 49.10 |
| | 5 | √ | √ | √ | √ | √ | 80.18 | 84.39 | 82.51 | 78.02 | 66.46 |
| G²CL | 1 | √ | | | √ | √ | 94.34 | 95.60 | 93.11 | 88.78 | 74.94 |
| | 2 | | √ | | √ | √ | 93.56 | 95.13 | 91.49 | 86.75 | 72.68 |
| | 3 | | | √ | √ | √ | 93.87 | 95.61 | 91.81 | 86.49 | 71.79 |
| | 4 | √ | √ | √ | | √ | 94.20 | 95.70 | 93.73 | 90.20 | 76.38 |
| | 5 | √ | √ | √ | √ | √ | 94.51 | 95.77 | 94.42 | 91.32 | 77.30 |
| | 6 | | | √ | | √ | 90.68 | 92.30 | 88.49 | 83.02 | 68.83 |

### 4.4 Ablation Study

In order to evaluate the contribution of each component in G²CL, a series of ablation experiments have been conducted on DenseUAV dataset. All experiments presented in this section utilise the setting(6) as mentioned in section 4.3.

In the baseline method, we use the shared parameter network as the backbone and training only with Eq.(1), and both training setting and inference produce are as same as G²CL. In the ablation experiments, we test the effectiveness of each component of $L_G$. For specifically, we test the contribution of $L_{GS}$, $L_{GU}$ and $L_{GC}$ by training combine with $L_p$, respectively. The result are shown in the Table 2.

As evidenced in Table 2, the performance of the system improves when each component is utilized independently. This serves to validate that each component contributes to the overall performance enhancements observed. Specifically, as observed in (2) and (3), the separate application of $L_{GS}$ and $L_{GU}$ resulted in improvements of 0.1% and 1.4% in Recall@1, respectively. It is important to note that although $L_{GC}$ did not yield a positive improvement when used alone, (5) and (6) demonstrate that combining $L_{GC}$ with $L_{GS}$ and $L_{GU}$ led to an approximately 2% increase in Recall@1. Moreover, a comparison between (5) and (8) further indicates that the addition of $L_{GC}$ can enhance Recall@1 by approximately 4%.

Additionally, it is noteworthy that $L_{GU}$ offers more significant improvements compared to $L_{GS}$. This is likely because, under the current setting (6), there is a greater number of images taken from different altitudes of the same location by UAVs, which increases the number of positive and negative contrast images to some extent. Similarly, as observed in (5) and (6) of Table 1, increasing the number of images of varying sizes from the satellite perspective also leads to a significant enhancement in model performance. As the number of positive and negative sample pairs increases, the model's performance improves significantly. This further demonstrates that the proposed loss function $L_G$ effectively captures the valuable information between positive and negative sample pairs, thereby validating the effectiveness of our proposed method.

Table 2. Ablation Study on DenseUAV under Setting(6) in Section 4.3

| Index | Component | | | Recall@1 | SDM@1 | SDM@3 | SDM@5 | SDM@10 |
|-------|-----------|---|---|----------|-------|-------|-------|--------|
| | $L_{GS}$ | $L_{GU}$ | $L_{GC}$ | | | | | |
| 1 | | | | 86.31 | 89.49 | 84.73 | 78.95 | 66.38 |
| 2 | √ | | | 86.40 | 89.47 | 84.77 | 78.80 | 65.75 |
| 3 | | √ | | 86.77 | 89.69 | 85.07 | 79.12 | 66.90 |
| 4 | | | √ | 84.25 | 87.84 | 82.63 | 77.33 | 65.06 |
| 5 | √ | √ | | 86.70 | 89.91 | 85.01 | 79.18 | 65.91 |
| 6 | | √ | √ | 88.15 | 90.55 | 86.55 | 81.17 | 67.65 |
| 7 | √ | | √ | 88.59 | 90.96 | 86.87 | 81.51 | 68.01 |
| 8 | √ | √ | √ | 90.68 | 92.30 | 88.49 | 83.02 | 68.83 |

### 4.5 More Evaluation and Analysis

In this section, we conduct an in-depth analysis of G²CL. Specifically, we perform detailed testing on the impact of the parameter $\alpha$ in Eq.(6) on G²CL performance. We also demonstrate the effects of different data settings on model performance and convergence speed. Finally, we visualize the ranking results and heatmaps obtained from satellite images using G²CL.

#### 4.5.1 Evaluation on Parameter $\alpha$

We conducted experiments to evaluate the impact of the parameter $\alpha$ in Eq. (6). Using the setting (6) from Section 4.3, the results are presented in Table 3. It can be observed that our model's performance is unaffected by variations in $\alpha$, with fluctuations remaining within 1% across the range of 1 to 5.

Table 3. Performance Comparison on different $\alpha$

| $\alpha$ | Recall @1 | SDM @1 | SDM @3 | SDM @5 | SDM @10 |
|----------|-----------|--------|--------|--------|---------|
| 1 | 90.35 | 92.38 | 88.34 | 82.88 | 69.01 |
| 2 | 90.61 | 92.10 | 88.42 | 83.19 | 68.25 |
| 3 | 90.68 | 92.30 | 88.49 | 83.02 | 68.83 |
| 4 | 90.69 | 92.26 | 88.34 | 82.74 | 68.56 |
| 5 | 90.68 | 92.27 | 88.32 | 82.75 | 68.57 |

#### 4.5.2 Compare with Other Loss function

Given that the UAV self-localization task is inherently a subset of image retrieval[23], this section compares our proposed geographic information adaptive loss $L_G$ method with traditional loss functions used in image retrieval tasks, such as face recognition [24] and person re-identification[25]. Specifically, we evaluate the performance of the standard triplet loss[26] and the triplet loss integrated with hard sample mining strategies[27] within the G²CL framework. The results, presented in Table 5, demonstrate that our proposed method achieves superior performance under the same data sampling strategy.

Table 4. Compare with different loss function, R@1 means Recall@1.

| Loss | R@1 | SDM@1 |
|------|-----|-------|
| Baseline | 86.31 | 89.49 |
| $L_p$ + Triplet Loss | 84.47 | 88.15 |
| $L_p$ + Hard Mining Triplet Loss | 86.53 | 89.82 |
| $L_G$ | 90.68 | 92.30 |

### 4.5.3 Visualization Results

In this section, we visualize the features generated by G²CL and compare them with those produced by state-of-the-art cross-view geo-localization method [12], as shown in Fig. 3. Our observations reveal that, in the context of UAV self-localization, our method prioritizes identifying the central visual position rather than focusing on matching typical targets. In contrast, cross-view geo-localization methods tend to concentrate more on typical targets, even when the viewpoint changes. This distinction aligns with the specific requirements of each task. Additionally, it is evident that as the viewpoint changes, the focus of features in UAV self-localization tasks shifts accordingly, further demonstrating that our method has a certain degree of spatial change awareness.

## 5. Discussion

The results of our study demonstrate that by leveraging geographic information, our approach not only enhances object localization but also improves the model's ability to discern fine-grained spatial differences, a crucial factor in precise UAV positioning. These findings underscore the importance of integrating geographic auxiliary data within contrastive learning frameworks to achieve more reliable and accurate localization. The ability to navigate with such precision is vital for the expanding range of UAV applications, particularly in scenarios where traditional localization methods fall short. As UAVs become increasingly integral to various industries, the implications of this research extend to improving the overall effectiveness and safety of autonomous operations.

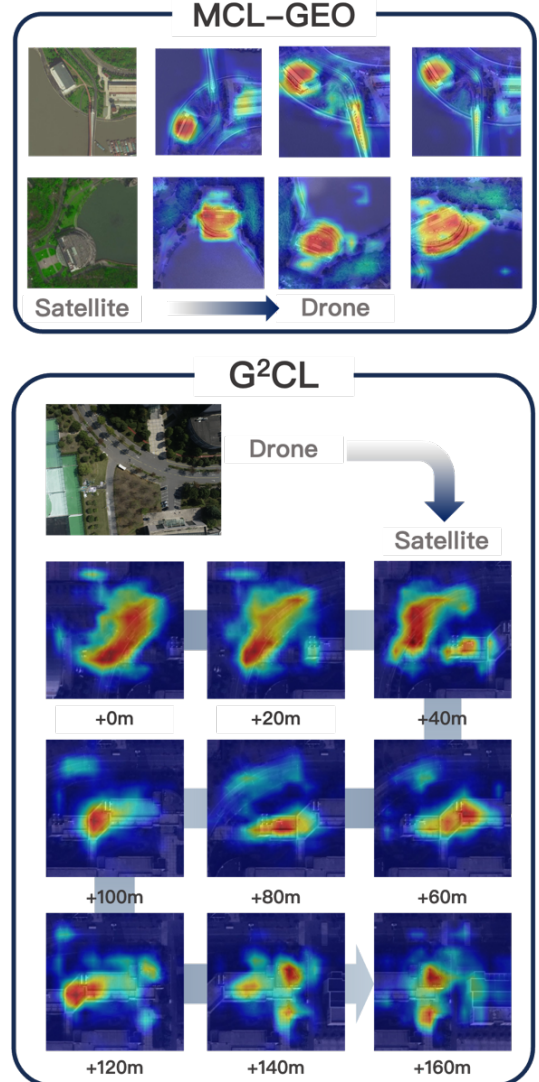

Fig. 3. The feature heatmap of MCL-GEO[12] and our proposed method. It can be found that for the task of cross-view geo-localization, the model pays more attention to the salient objects in the field of view, while for the UAV self-positioning task, the model pays more attention to feature extraction in the central field of view, which also changes as the perspective advances.

## 6. Conclusion

In this paper, we focus on addressing the critical challenge of UAV self-positioning in environments where GNSS signals are unavailable or unreliable. To overcome the limitations of traditional cross-view geo-localization methods, which often neglect spatial discrepancies, we proposed a novel approach called Geographic Information Guided Contrastive Learning (G²CL). By integrating geographic auxiliary information into a contrastive learning framework, our method not only available to locate objects but also distinguish fine-grained distance differences. In future work, we will explore the computational limitations of UAVs and focus on developing more efficient real-time processing algorithms to enable vision-based autonomous UAV localization.